\documentclass[10pt,twocolumn,letterpaper]{article}

\usepackage[pagenumbers]{cvpr} 

\usepackage[dvipsnames]{xcolor}

\definecolor{cvprblue}{rgb}{0.21,0.49,0.74}
\usepackage[pagebackref,breaklinks,colorlinks,citecolor=cvprblue]{hyperref}
\usepackage{soul}
\usepackage{color, xcolor}
\usepackage{upgreek}
\usepackage{amsmath,amsthm,amssymb,amsfonts}
\usepackage{svg}
\usepackage{stfloats}
\usepackage{indentfirst}
\usepackage{multirow}
\usepackage{multicol}
\usepackage{makecell}
\usepackage{setspace}

\def\paperID{16920} 
\def\confName{CVPR}
\def\confYear{2024}

\title{Depth-aware Volume Attention for Texture-less Stereo Matching}

\author{
Tong Zhao$^{1}$ \quad Mingyu Ding$^{2}$ \quad Wei Zhan$^{2}$ \quad Masayoshi Tomizuka$^{2}$ \quad Yintao Wei$^{1}$\\
$^{1}$Tsinghua University \quad $^{2}$UC Berkeley
}

\begin{document}
\maketitle
\begin{abstract}

Stereo matching plays a crucial role in 3D perception and scenario understanding. Despite the proliferation of promising methods, addressing texture-less and texture-repetitive conditions remains challenging due to the insufficient availability of rich geometric and semantic information. In this paper, we propose a lightweight volume refinement scheme to tackle the texture deterioration in practical outdoor scenarios. Specifically, we introduce a depth volume supervised by the ground-truth depth map, capturing the relative hierarchy of image texture. Subsequently, the disparity discrepancy volume undergoes hierarchical filtering through the incorporation of depth-aware hierarchy attention and target-aware disparity attention modules. Local fine structure and context are emphasized to mitigate ambiguity and redundancy during volume aggregation. Furthermore, we propose a more rigorous evaluation metric that considers depth-wise relative error, providing comprehensive evaluations for universal stereo matching and depth estimation models. We extensively validate the superiority of our proposed methods on public datasets. Results demonstrate that our model achieves state-of-the-art performance, particularly excelling in scenarios with texture-less images. The code is available at \href{https://github.com/ztsrxh/DVANet}{https://github.com/ztsrxh/DVANet}

\end{abstract}
\section{Introduction}
\label{sec:intro}

Stereo matching, a fundamental task in 3D vision, holds significant applications in scenario reconstruction and understanding \cite{review_PAMI}\cite{9394777}. Its widespread utility spans areas such as augmented reality, robotics, and perception in autonomous driving. Stereo matching generally consists of four cascaded modules: image feature extraction, cost volume construction, cost aggregation and disparity estimation \cite{988771}. Deep learning has significantly promoted its progress and achieved remarkable performance.

\begin{figure}[t]
  \centering
   \includegraphics[width=0.99\linewidth]{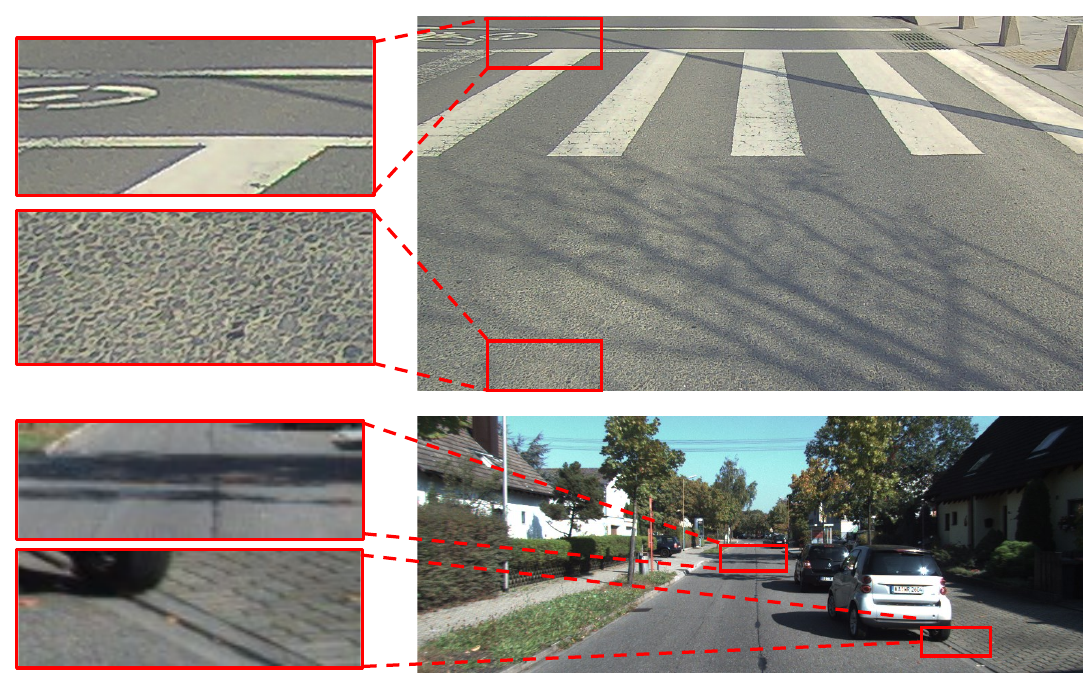}
   \vspace{-8pt}
   \caption{\textbf{Our motivation.} The perspective effect leads to texture deterioration in natural scenarios. Texture is relatively rich at small depth, while degenerates at farther distance. }
   \label{fig:texture_example}
\end{figure}

\begin{figure}[t]
  \centering
   \includegraphics[width=0.99\linewidth]{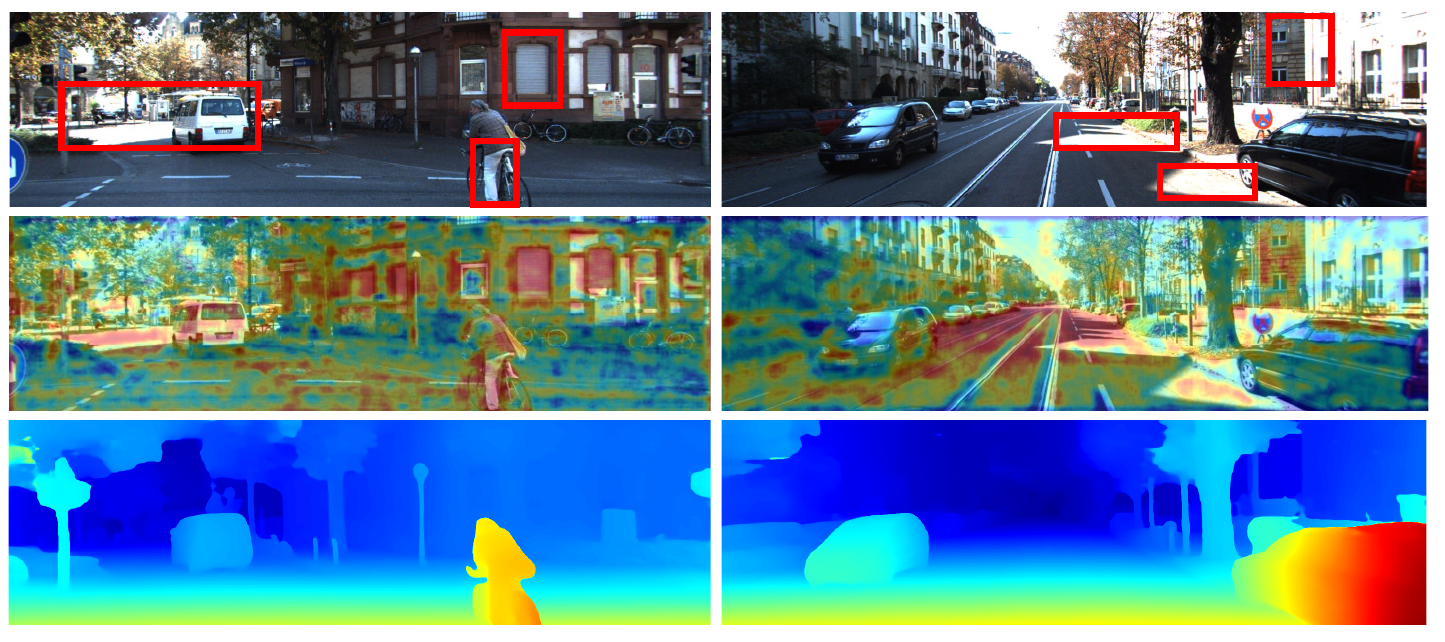}
   \vspace{-8pt}
   \caption{Texture attention. From up to down: left image, texture hierarchy attention map, and predicted disparity map. Our DVANet concentrates on the \textbf{texture-less and weak-texture} areas such as the over-exposed regions, white windows, and roads at far distance in the images. }
   \vspace{-8pt}
   \label{fig:attention_example}
\end{figure}

Despite the emergence of advanced methods such as transformer-based \cite{CSTR}\cite{DLNR}\cite{ELFNet}\cite{STTR} and iterative-based \cite{IGEV}\cite{raft}\cite{DLNR} models, there are still performance limitations particularly in handling large texture-less and texture-repetitive areas such as road surface. The texture sparseness and deterioration led by the perspective effect in natural scenarios are generally neglected in existing works.  As shown in Figure \ref{fig:texture_example}, at near distance, a pixel corresponds to small physical scale and image texture is distinct. However, as the depth increases, a single pixel covers a larger physical area, leading to a gradual deterioration of texture definition. Texture details of objects are lost at farther distance. This texture hierarchy property linked to depth poses significant challenges in extracting discriminative local features especially at larger depth. 

The cost volume construction and aggregation are essential steps determining the comprehensive performance of disparity regression \cite{9157353}\cite{8578395}\cite{Yang_2020_CVPR}\cite{rs10111844}. The 4D similarity volume is generally derived by the non-parametric correlation operation \cite{DispNet}. Cascaded 3D convolutions are then employed to aggregate the similarity information, which demands much computation and memory during training \cite{8953478}. The subsequent works focus on exploring more effective volume construction and aggregation with less 3D convolutions. Examples include GwcNet \cite{GwcNet} and ACVNet \cite{ACVNet}, which leverage group-wise correlation and volume attention to regularize the similarity volume.
Nevertheless, these operations are not adaptive to the feature patterns of weak texture, remaining the potential problem of information loss or redundancy. The feature discrepancy between the two images, which originally should be accented for the benefit of texture-less matching, is instead suppressed by the correlation or concatenation. Moreover, the volume construction is far away from the ground-truth supervision, leaving difficulties in learning effective similarity representations. 

In this paper, we concentrate on efficient stereo matching in severely texture-less scenarios. Motivated by extensive insights into the texture patterns in natural images, we propose a straightforward yet promising model called Depth-aware Volume Attention Network (DVANet) to address the above limitations. A lightweight depth estimation branch is introduced at the low-level feature map to capture the geometric and structural hierarchies. The derived depth volume, with channels encoding relative depth distribution, serves as attention weight to assign more channel importance to texture features at larger depth. The texture weakening especially at farther distance caused by perspective effect in the image domain is thereby compensated in the feature domain. Jointly supervised by depth and disparity labels, the feature extraction and volume aggregation backbones concentrate on both global texture hierarchy and local texture details, as illustrated in Figure \ref{fig:attention_example}. Furthermore, to refine the aggregated volume and achieve more accurate disparity estimation, we design another disparity attention module to focus solely on the channel features near the target disparity value. 

Moreover, the existing works evaluate matching performance with metrics like end point error (EPE, the average disparity absolute error), and n-pixel error (the percentage of pixels with errors larger than n pixels) \cite{6248074} \cite{DispNet}. However, disparity describes only the coordinate difference of pixel correspondence in the image domain, offering a limited perspective that may not be directly applicable in the 3D physical world. The disparity error provides a coarse and overall evaluation of model performance without considering depth-wise error. Given that stereo matching aims to recover the 3D scenario, it becomes imperative to introduce a metric assessed in the physical phase. To bridge this gap, we propose the Weighted Relative Depth Error (WRDE), a more comprehensive evaluation metric applicable to both stereo matching and depth estimation methods. 

Our main contributions are summarized as follows. 1) We design a depth-aware texture hierarchy attention module and a target-aware disparity attention module tailored for texture-less stereo matching. 2) We propose a novel and universally applicable evaluation metric for depth estimation methods, encompassing monocular depth, stereo matching, and multi-view stereo. 3) Through comprehensive experiments, we demonstrate the effectiveness of our methods in enhancing performance on public datasets.

\section{Related Works}
\label{sec:related}

\noindent {\bf Cost volume and aggregation.} Cost-volume based stereo matching has achieved remarkable performance \cite{9009458}\cite{8237279} \cite{song2019edgestereo}\cite{review_PAMI}. DispNetC \cite{DispNet} is the first to leverage feature map correlation to create a 3D volume, enabling end-to-end deep stereo matching. To tackle the limited information by the full correlation, GC-Net \cite{GCNet} concatenates the features along the channel dimension to construct a 4D volume. However, due to the absence of direct encoding of matching similarity, more 3D convolutions are required for effective volume aggregation. To balance informative representation and computation consumption, GwcNet \cite{GwcNet} introduces a combination of group-wise correlation and concatenation to form a composite volume. However, there may be an information gap in direct concatenation without considering their own distributions \cite{ACVNet}. 
 
To enhance the advantages of the combined volume, ACVNet \cite{ACVNet} filters the concatenation volume using another parallel correlation volume. The attention module directs subsequent aggregations to focus exclusively on similarities near the actual disparity values. Nevertheless, the sparse attention weights are peaked only at the target disparities due to the strong sifting property of the \emph{soft-max} operation. Consequently, a significant portion of similarity measures is disregarded at the initial volume without utilization, posing challenges in volume aggregations. PSMNet \cite{PSMNet} designs serial hourglass modules each supervised by the same disparity map to regularize the cost volume. The repetitive encoder and decoder operations provide almost the same gradient information, bringing fractional improvements to disparity estimation accuracy. Cascaded volumes, such as CFNet \cite{CFNet} and CasStereo \cite{CasStereo} are proposed to establish a volume pyramid at different resolutions and estimate the final disparity in a coarse-to-fine manner. However, the limited representation capability of low-resolution volumes introduces errors that propagate and accumulate in later stages. 


\noindent {\bf Processing of texture-less images.} 	Stereo matching in texture-less and weak-texture regions is constantly a teaser. Many efforts have been directed toward enhancing feature extraction backbones. CSTR \cite{CSTR} and STTR \cite{STTR} demonstrate the usage of transformers in extracting long-range contextual and positional information such as object orientation and edge information to enhance the representation capability in texture-less areas. However, the semantic context in natural scenarios is often complex, leading to models with limited robustness against various weak texture patterns. Concentrating on the local texture variation is more promising than capturing global scenario representation.

Some works aim to alleviate matching ambiguity by constraining the similarity distribution in the cost volume. AcfNet \cite{AcfNet} imposes an unimodal constraint on the output volume, leading to a concentrated probability distribution around the actual disparity. CFNet \cite{CFNet} introduces uncertainty estimation to quantify the degree of multimodal distribution in the volume, thereby adjusting the disparity search range in the coarse-to-fine estimation scheme. Despite achieving impressive performance, the non-single probability assignment still causes ambiguity on the adjacent disparities.

\section{Methods}

\begin{figure*}[ht]
\setlength{\abovecaptionskip}{0.3cm}
\setlength{\belowcaptionskip}{-0.2cm}
  \centering
   \includegraphics[width=0.99\linewidth]{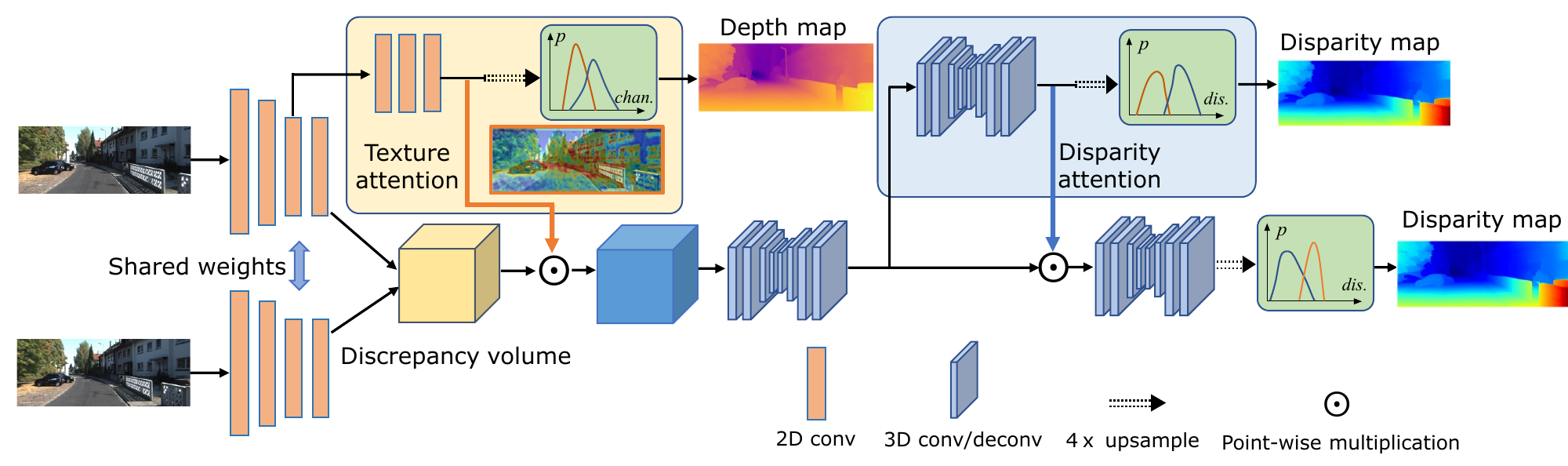}
   \vspace{-6pt}
   \caption{The architecture of the proposed DVANet. It contains three novel designs: discrepancy volume adapted to texture-less matching, depth-aware texture hierarchy attention, and target-aware disparity attention. The network is guided to focus on the relative texture hierarchy thus achieving more reliable matching in weak-texture areas.}
   \label{fig:framework}
\end{figure*}

To address the identified challenges, we propose our DVANet (see Figure \ref{fig:framework}) for effective texture-less stereo matching. In this section, we first introduce the evaluation metric WRDE for depth estimation models, which simultaneously explains our motivation of designing the methods. Then, we give detailed descriptions about the model architecture.

\subsection{Weighted Relative Depth Error} \label{Weighted Relative Depth Error}

The widely utilized metric EPE in stereo matching cannot provide insightful descriptions of model performance. For binocular stereo vision, the relationship between disparity \emph{d} and depth \emph{z} is described as:
\begin{spacing}{1}
\begin{equation}
  d=\frac{fb}{z}
  \label{eq:FBZ}
\end{equation}
\end{spacing}

 \noindent where \emph{f} and \emph{b} are the focal length and stereo baseline, respectively. Assuming that there is an error $e_d$  between the ground-truth disparity $d_{gt}$ and the estimated disparity \emph{\^{d}}, i.e.:
\begin{equation}
  \hat{d}=d_{gt}+e_d
  \label{eq:dis_err}
\end{equation}

We can derive the relative depth error $\overline{e_z}$ combining with Equation \ref{eq:FBZ}:
\begin{equation}
  \overline{e_z}=\frac{\left|z_{gt}-\hat{z}\right|}{z_{gt}}=\left|1-\frac{d_{gt}}{\hat{d}}\right|=\frac{1}{\left|1+\frac{d_{gt}}{e_d}\right|}.
  \label{eq:dep_abs_err}
\end{equation}

Equation \ref{eq:dep_abs_err} illustrates that a constant EPE value corresponds to varying depth errors at different depths, which increase nonlinearly at farther distances. Given that stereo matching models often exhibit varied performance at distinct depths, equivalent EPE values for two models do not necessarily imply comparable overall performance. Metrics like EPE or the percentage of outliers may not offer a holistic assessment of stereo matching performance.

\begin{figure}[t]
\setlength{\abovecaptionskip}{0.3cm}
\setlength{\belowcaptionskip}{-0.2cm}
  \centering
   \includegraphics[width=0.9\linewidth]{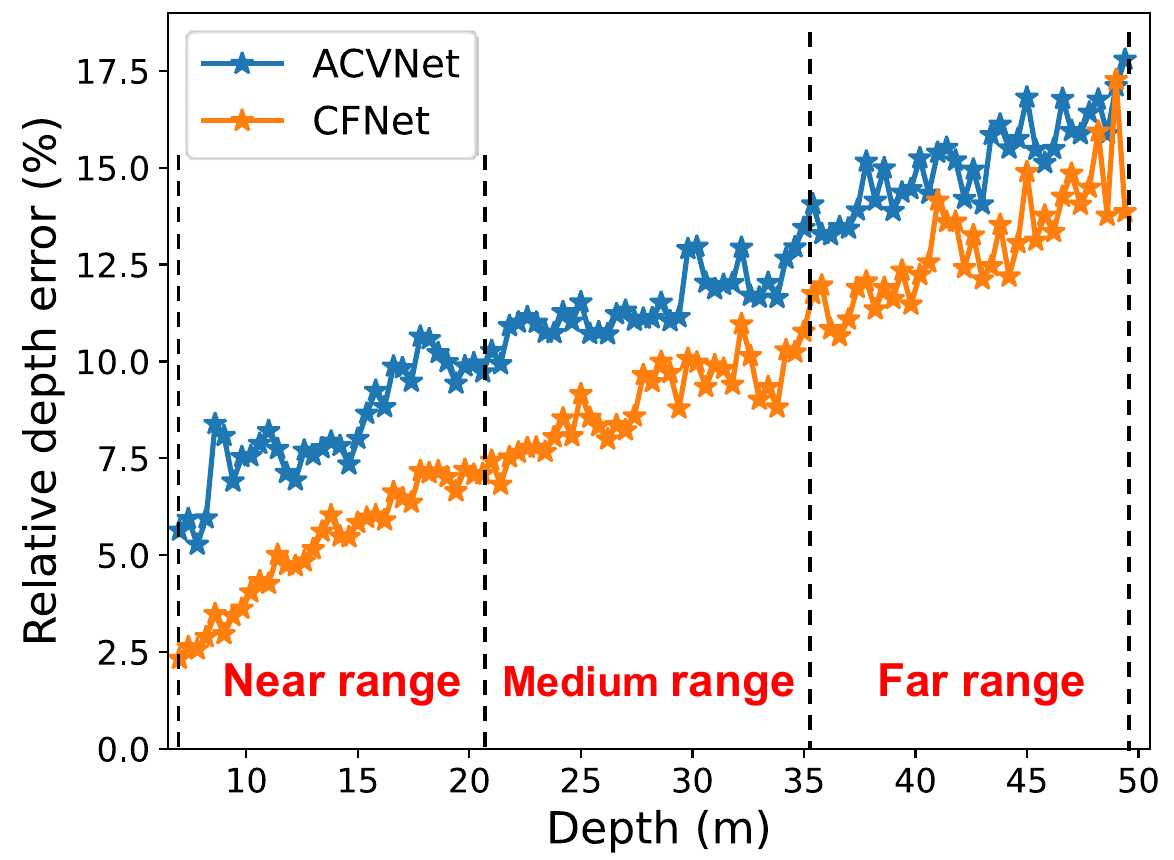}
   \vspace{-6pt}
   \caption{Visualization of relative depth error w.r.t. depth. Results are from the cross-domain generalization evaluation on the KITTI 2015. The errors increase with depth, and the models exhibit distinct performance.}
   \label{fig:wrde_example}
\end{figure}

To experimentally reveal the practical relationship between $\overline{e_z}$ and depth \emph{z}, we visualize the results from the cross-domain generalization of ACVNet \cite{ACVNet} and CFNet \cite{CFNet} on the train set of KITTI 2015 \cite{Menze2015CVPR}, as presented in Figure \ref{fig:wrde_example}. We set the depth range of interest as 7$\sim$50 m with a depth interval of 0.4 m. The relative depth errors of points within the intervals are averaged. The resulting curves highlight that the error tends to increase with depth, aligning with our theoretical analysis above. The performance decrease of practical matching at far distance is originated from the ambiguity in weak-texture features, which is further caused by the hierarchy property from perspective effect. 

For stereo matching in natural and outdoor scenarios, errors at larger depths are dominate. A model that exhibits higher accuracy at greater distances is inherently superior. Hence, we propose the WRDE to meticulously assess the depth-wise performance. To calculate the metric, the depth range of interest and depth resolution are first set based on the statistical characteristics of the target dataset. The entire depth range is then equally divided into three segments: near, medium, and far ranges. To emphasize their distinctive importance, we assign relative weights of 1, 2, and 3 to these segments, respectively. The final WRDE value is derived through the normalized weighted sum of all error samples:
\begin{equation}
    WRDE=\sum_{i=1}^{N}{w_i\overline{e_{zi}}},\\
    w_{i} = \left\{ \begin{matrix}
    {\frac{1}{2N},~1 \leq i < \frac{N}{3}} \\
    {\frac{1}{N},~\frac{N}{3} \leq i < \frac{2N}{3}} \\
    {\frac{3}{2N},~\frac{2N}{3} \leq i \leq N}
    \end{matrix} \right.
  \label{eq:wrde}
\end{equation}

\noindent where \emph{N} is the number of total errors, and $w_{i}$ is the weight of the sample. The analysis method and metric are also applicable to other 3D vision tasks aiming at depth recovery such as monocular depth estimation and multi-view stereo.

\subsection{Network Architecture} \label{Network Architecture}

The structure of the proposed DVANet, as illustrated in Figure \ref{fig:framework}, incorporates three key innovations: discrepancy volume construction, depth-aware hierarchy attention, and target-aware disparity attention modules. Despite the superior long-range dependency modeling capabilities of transformer-based backbones \cite{liu2021Swin}\cite{lin2022survey}, we opt for the simplified EfficientNet \cite{tan2019efficientnet} for feature extraction. This choice aligns with prior knowledge of stereo vision constraints, where we focus solely on local feature similarities along the same rows.  For images with weak texture, pixel serialization would corrupt feature extraction on fine details and introduce unnecessary ambiguities. Thus, we emphasize the focus on slight local texture variation over global but coarse scenario encoding. We anticipate that CNN-based models, coupled with appropriate regularization strategies, will demonstrate improved performance for texture-less images.

To preserve more local fine texture structures, we set larger kernel size of 9, 7, and 5 at the initial three convolutions. We stack four MBConv blocks in the backbone. Unlike the original architecture, we down-sample the feature maps at later stages for extracting richer geometric and texture features. The feature maps $\emph{\textbf{F}}_l$ and $\emph{\textbf{F}}_r$ with the shape of [128, \emph{H}/4, \emph{W}/4] are derived, where \emph{H} and \emph{W} are the input image height and width, and 128 is the number of channels.

To accentuate discriminative information from analogous features of texture-less images, we adopt the subtraction operation instead of correlation or concatenation when constructing the 4D cost volume:
\begin{equation}
    \mathbf{C}_{discr}\left(\cdot,\ d,\ x,\ y\right)=\boldsymbol{F}_\mathit{l}\left(x,y\right)-\boldsymbol{F}_\mathit{r}(x-d,\ y)
  \label{eq:volume}
\end{equation}

The discrepancy volume above emphasizes subtle difference between similar features, contributing to effective cost measure. To reduce computational burden and achieve a more compact representation, we reduce channel dimension by group-wise average \cite{GwcNet}. The size of the reduced discrepancy volume is [32, $\emph{d}_{max}/4$, \emph{H}/4, \emph{W}/4], where $\emph{d}_{max}$ is the maximum disparity. The volume is then filtered by the proposed depth-aware hierarchy attention module, which will be described in Section \ref{Depth-aware Hierarchy Attention}. 

 Aggregations by combined 3D convolutions and hourglass structures are implemented to refine the similarity features. Unlike the previous works which further regularize the volume with serial encoder-decoder predictions supervised by disparity maps, we introduce a target-aware disparity attention module at the output stage. The detailed structure will be introduced in Section \ref{Target-aware Disparity Attention}. The predicted disparity probability volume is at 1/4 full resolution, which undergoes cascaded interpolation, \emph{soft-max}, and \emph{soft argmin} operations to produce the final disparity map $\boldsymbol{D}_{1}$.

\subsection{Depth-aware Hierarchy Attention} \label{Depth-aware Hierarchy Attention}

Building upon the aforementioned insights gained from image patterns in practical natural scenarios, we recognize the importance of texture hierarchy induced by the perspective effect in influencing matching performance. To leverage this instructive prior knowledge, we devise an attention module that encodes depth-related texture hierarchy to filter pixel features along the channel dimension. Specifically, we introduce a lightweight depth estimation task branching from the early feature extraction stage on the left image. It effects by concentrating on the low-level feature maps that preserve informative geometry and structure contexts. In contrast to general monocular depth estimation methods employing deep encoder-decoder structures \cite{PSMNet}\cite{GwcNet}, we use three cascaded MBConv blocks at the same resolution, resulting in the feature map $\emph{\textbf{F}}_{att}$ with a size of [32, $\emph{H}/4$, $\emph{W}/4$], encoding relative depth and hierarchy information.

In typical depth estimation methods, $\emph{\textbf{F}}_{att}$ is usually reduced to a single-channel map and exported as relative depth using the \emph{sigmoid} function \cite{LaDepth}\cite{idisc}\cite{AdaBins}. However, drawing inspiration from the disparity probability volume, we treat the map as a depth volume, with each channel representing corresponding depth probabilities. For end-to-end learning, we obtain the full-resolution depth map with the similar style of disparity regression. The normalized depth map prediction $\overline{\boldsymbol{Z}_{p}}$ is calculated as:

\begin{equation}
    \overline{\boldsymbol{Z}_{p}}(x,y)=\sum_{c=0}^{31}{\overline{z_c}\cdot Softmax({{\boldsymbol{F}_{att}}^\prime(\cdot,\ x,\ y)})}
  \label{eq:depth_pred}
\end{equation}

\noindent where $\boldsymbol{F}_{att}^\prime$ is the interpolated full-resolution feature map, and $\overline{z_c}$=\emph{c}/31 is the relative depth index of channel \emph{c}. The depth-aware channel attention weights $\emph{\textbf{A}}_c$ are derived as:

\begin{equation}
    \boldsymbol{A}_\mathit{c}=\ \frac{1}{1+e^{-{\boldsymbol{F}_{\mathit{att}}} }}.
  \label{eq:channel_attention_weights}
\end{equation}

For normalization, we employ the \emph{sigmoid} function instead of \emph{soft-max} to achieve a gentle filter and preserve more information. The 4D discrepancy volume is then filtered by the pixel-wise attention weights:
\begin{equation}   
\boldsymbol{C}_{dha}\left(d\right)=\boldsymbol{A}_\mathit{c}\odot\boldsymbol{C}_{discr}(d)
  \label{eq:channel_filter}
\end{equation}

\noindent where $\odot$ represents the element-wise product broadcast to the disparity dimension. Guided by effective depth estimation, a pixel at near distance corresponds to a depth probability distribution peaked at the small channel index of $\boldsymbol{A}_c$, signifying higher attention weights of the nearby channels. This attention supervision enables the feature extractor to learn more discriminative representations, as the channel dimension is imbued with the significance of texture hierarchies.

The end-to-end learning of depth volume is supervised by the normalized depth map $\overline{\boldsymbol{Z}_{gt}}$, which can be directly converted from the ground-truth disparity map $\boldsymbol{D}_{gt}$:
\begin{equation}   
    \overline{\boldsymbol{Z}_{gt}}=\frac{d_{min}}{\boldsymbol{D}_{gt}}
  \label{eq:norm_depth_map}
\end{equation}
 where $\emph{d}_{min}$ is the minimum disparity of the dataset.

\subsection{Target-aware Disparity Attention} \label{Target-aware Disparity Attention}

The discrepancy volume with channel hierarchy attention is then aggregated by the 3D convolutions. Despite their strong representation abilities, information redundancy remains unavoidable as convolutions span the channel dimensions of all disparities. To address this, we introduce the target-aware disparity attention module, which focuses solely on similarity features near the actual disparity. In contrast to ACVNet \cite{ACVNet}, which applies the disparity filter soon after volume construction, we conduct attention weighting at a later stage. This approach enables both informative aggregation at the initial stage and refined disparity regression at the output stage.

We implement a concise aggregation and regression branch, comprising four 3D convolutions and one hourglass module applied to the aggregated volume $\boldsymbol{C}_{agg}$. The disparity probability volume $\boldsymbol{A}_d$, serving as the target-aware disparity attention weight, is obtained by applying \emph{soft-max} across the disparity dimension of the single-channel volume. The prior disparity estimation $\boldsymbol{D}_0$ is calculated in the same style as $\boldsymbol{D}_1$. The volume $\boldsymbol{C}_{agg}$ is then filtered by element-wise product broadcast to the channel dimension:
\begin{equation}   
    \boldsymbol{C}_{tda}\left(c\right)=\boldsymbol{A}_\mathit{d}\odot\boldsymbol{C}_{agg}(c)
  \label{eq:dis_attention}
\end{equation}

The disparity probability volume emphasizes the similarity features near the target value, providing beneficial constraint to the later aggregations. It contributes to more effective regularization with less expensive 3D convolutions. 

\subsection{Loss Function} \label{loss function}
We calculate the smooth L1 loss to the prior and formal disparity predictions:

\begin{equation}   
    \mathcal{L}_{{D}_0,{D}_1}={Smooth}_{L_1}(\boldsymbol{D}_{0,1}-\boldsymbol{D}_{gt})
  \label{eq:dis_loss}
\end{equation}

Unlike the depth estimation methods which generally adopt the scale-invariant loss \cite{lee2019big}\cite{LaDepth}\cite{monodepth17}, we also calculate the smooth L1 loss for the normalized depth map.


To avoid the complicated parameter fine-tuning that weights the three losses, we normalize the scale of every loss thus adaptively balancing the gradients: 
\begin{equation}   
\mathcal{L}_{total}=\overline{\mathcal{L}_{D_0}}+\overline{\mathcal{L}_{D_1}}+\overline{\mathcal{L}_{dep}}
  \label{eq:total_loss}
\end{equation}
where $\overline{\mathcal{L}_{dep}}$ is the normalized depth loss.

\section{Experiments}

In this section, we comprehensively validate the performance of our DVANet on public evaluation benchmarks including both synthetic and real-world datasets. Ablation analyses are performed to verify the effectiveness of the proposed modules. Additionally, numerical and visualized results are also presented to demonstrate the usability and superiority of WRDE. 

\subsection{Datasets and Metrics}

\textbf{Scene Flow} \cite{DispNet} is a large-scale synthetic dataset consists of 35,454 training and 4,370 testing stereo pairs. As the images are synthetic, it has large texture-less areas in both the background and foreground objects. We adopt the finalpass subset for pre-training the model. 

\textbf{KITTI 2012} and \textbf{KITTI 2015} \cite{6248074}\cite{Menze2015CVPR}  are real-world datasets for autonomous driving perception. The former has 194 training and 195 testing image pairs, and the latter contains 200 pairs in both the training and testing sets. Since the image resolution is only at 1242×375, much fine texture in the large-scale natural scenarios is lost. Additionally, there are many over-exposed areas due to the poor dynamic range of cameras.

\textbf{RSRD} \cite{rsrd} is a real-scenario and high-resolution dataset for road surface reconstruction purpose, providing dense disparity labels acquired by LiDAR. The dataset comprises 2,500 training and 300 testing image pairs with a resolution of 1920×1080 in the dense subset. It is challenging for stereo matching algorithms as the texture of road surface is inherently in lack without certain patterns. We demonstrate the performance on this dataset as our DVANet is specifically designed for stereo matching in severely texture-less and weak-texture scenarios.

Following the common evaluation methods, we adopt EPE, \textgreater 1px or \textgreater 2px errors (percentage of pixels with error larger than 1 or 2 pixels), and D1-3px (percentage of outlier pixels) for assessing the model performance. We implement the proposed WRDE metric only on the real-world datasets since the depth range in the Scene Flow dataset is generally discontinuous. The depth range of interest is set as 7$\sim50$ m with interval of 0.40 m for KITTI datasets, while 2$\sim$8 m with interval of 0.15 m for RSRD.

\begin{figure*}[t]
\setlength{\abovecaptionskip}{0.3cm}
\setlength{\belowcaptionskip}{-0.2cm}
  \centering
   \includegraphics[width=1\linewidth]{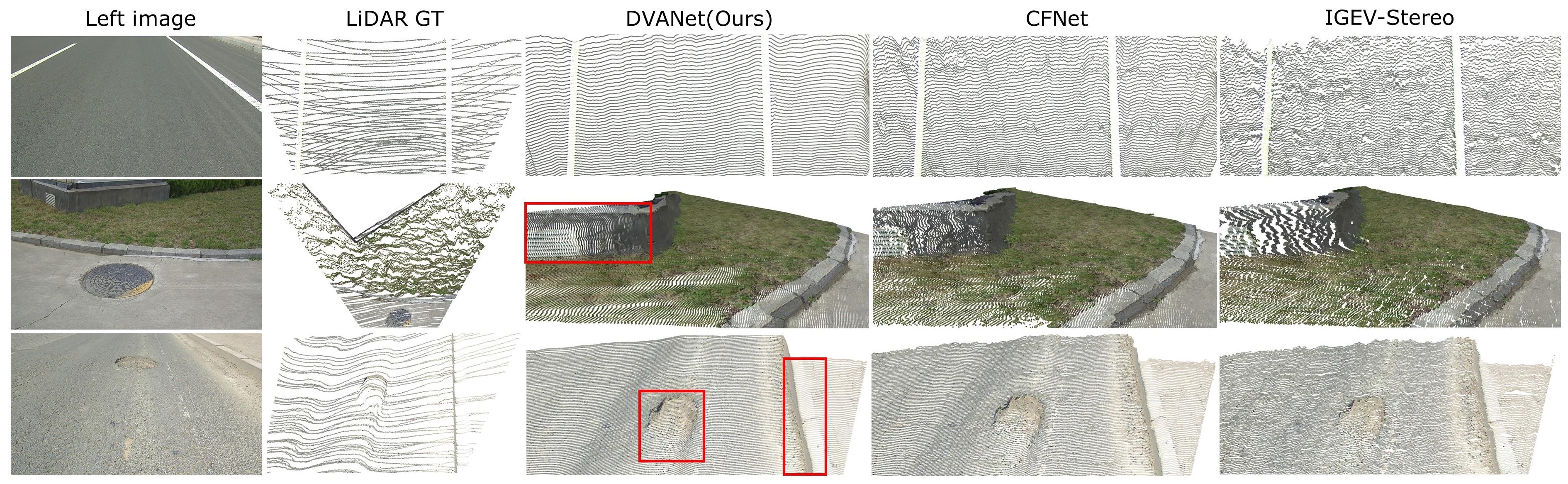}
   \caption{Visualization of point clouds converted from disparity maps on RSRD. Our DVANet delivers accurate and stable disparity estimation on the extremely texture-less plan road. Fine local structures are accurately recovered even at far distance. 
   }
   \label{fig:rsrd_visualization}
\end{figure*}

\subsection{Implementation Details} \label{Implementation Details}

We implement our network with the PyTorch framework and conduct experiments with single NVIDIA A100 GPU. For all runs, we use the AdamW optimizer with weight decay of 1e-4. We adopt the OneCycle learning rate scheduler with the linear anneal strategy. The maximum learning rate is set as 0.0014. The original image resolution of RSRD is too big for model training, so we resize and crop the image to 512×960, which is the same for the Scene Flow dataset. The maximum disparity $d_{max}$=192 for Scene Flow and KITTI datasets, while 64 for RSRD. For computing the normalized depth map, we set the minimum disparity $d_{min}$ as 1.0 and 3.8 for the Scene Flow and KITTI datasets, respectively. Since the RSRD provides ground-truth depth, we set the maximum depth as 13.0 m to directly normalize the map. The training images are directly fed into the network without any augmentation. 

\subsection{Comparison with State-of-the-art} \label{sota_comparison}

\begin{table}[t]
  \centering
  \caption{Comparison with state-of-the-art on RSRD. \textbf{Bold}: best. \underline{Underline}: second best.}
  \resizebox{\columnwidth}{!}{%
    \begin{tabular}{l|c|c|c|c}
    \hline
    Method &  \makecell[c]{EPE \\ (px)} & \makecell[c]{\textgreater 1px \\ (\%)} & \makecell[c]{\textgreater 2px \\ (\%)} & \makecell[c]{WRDE \\ (\%)} \\
    \hline
    IGEV-Stereo \cite{IGEV} & 0.19 & 0.54 & 0.22 & 0.81 \\
    CFNet \cite{CFNet} & 0.18 & 0.90 & 0.17 & 0.74  \\
    PSMNet \cite{PSMNet} & 0.17 & 0.63 & 0.16 & 0.71  \\
    RAFT-Stereo \cite{raft} & 0.17 & \underline{0.43} & 0.17 & 0.72  \\
    ACVNet \cite{ACVNet} & \underline{0.16} & 0.59 & \underline{0.15} & \underline{0.69} \\
    GwcNet \cite{GwcNet} & \underline{0.16} & 0.57 & \textbf{0.14} & \underline{0.69}  \\
    \hline
    DVANet (Ours) & \textbf{0.15} & \textbf{0.34} & \textbf{0.14} & \textbf{0.62} \\
    \hline
    \end{tabular}%
  }
  \label{tab:sota_rsrd}%
\end{table}%

\textbf{RSRD.} For fair comparison, we retrain the models in Table \ref{tab:sota_rsrd} on RSRD with the provided configurations without any modifications and then test their performance respectively. All the models are trained for 5 epochs with batch size of 2. As demonstrated in Table \ref{tab:sota_rsrd}, our DVANet comprehensively outperforms the other methods on all the metrics. Our method improves 1-pixel error rate by 20.9\% compared with the RAFT-Stereo \cite{raft}, and surpasses the second by 10.1\% as for the WRDE. 

For further comparison, Figure \ref{fig:rsrd_visualization} visualizes point clouds generated from disparity maps using the provided extrinsic parameters in RSRD \cite{rsrd}. In the first scenario, representing a flat road without obvious undulations, our DVANet exhibits continuous scanlines (i.e., pixels on the same row) with minimal longitudinal vibrations, indicating accurate and stable disparity estimation. In contrast, the scanlines derived from IGEV-Stereo \cite{IGEV} are severely intermittent, introducing much noise on the flat road surface.  The other two scenarios also demonstrate the superior performance of our method in handling large texture-less areas. The fine surface structures at far distance are accurately recovered such as the wall, pothole, and road side stone.

\textbf{Scene Flow.} We first train the model on the full set for 20 epochs and then train on the FlyingThings3D subset for 5 epochs with the maximum learning rate at 5e-4. As shown in Table \ref{tab:sota_sceneflow}, our model ranks best as for the 1-pixel error rate, outperforming the transformer-based and iterative-based models such as DLNR \cite{DLNR} and RAFT-Stereo \cite{raft}. Compared with the state-of-the-art method IGEV-Stereo \cite{IGEV}, our DVANet improves the 1-pixel error rate and D1-3px by 3.6\% and 19.8\%, respectively. 

\begin{table}[t]
  \centering
  \caption{Comparison with state-of-the-art on Scene Flow. The disparity range is 0$\sim$192. \textbf{Bold}: best. \underline{Underline}: second best.}
    \begin{tabular}{p{7.2em}|c|c|c}
    \hline
    Method & EPE(px) & \textgreater 1px(\%) & D1-3px(\%)  \\
    \hline
  
    PSMNet\cite{PSMNet} & 1.09 & 12.10 & {-}  \\
    CFNet\cite{CFNet} & 0.97 & 9.90 & 4.70  \\
    GANet-15\cite{GANet} & 0.84 & 9.90 & 4.52  \\
    LEAStereo\cite{LEAStereo} & 0.78 & 7.82 & {-} \\
    GwcNet\cite{GwcNet} & 0.76 & 8.03 & 2.71  \\
    RAFT-Stereo\cite{raft} & 0.65 & 7.32 & 2.93 \\
    DLNR\cite{DLNR} & \underline{0.48} & 5.38 & 2.36\\
    IGEV-Stereo\cite{IGEV} & \textbf{0.47} & 5.33 & 2.22  \\
    ACVNet\cite{ACVNet} & \underline{0.48} & 5.34 & \textbf{1.59}  \\
    \hline
    DVANet(Ours) & 0.53 & \textbf{5.14} & \underline{1.78} \\
    \hline
    \end{tabular}%
  \label{tab:sota_sceneflow}%
\end{table}%

\begin{table}[t]
  \centering
  \caption{Comparison with state-of-the-art on KITTI 2012. The metrics are evaluated based on the 3-pixel error. \textbf{Bold}: best. \underline{Underline}: second best.}
    \begin{tabular}{p{7.5em}|c|c|c}
    \hline
    Method &  Noc(\%) & All(\%) & Params(M) \\
    \hline
    HITNet\cite{HITNet} & 1.41 & 1.89 & {-} \\
    GANet-15\cite{GANet} & 1.36 & 1.80 & {-} \\
    RAFT-Stereo\cite{raft} & 1.30 & 1.66 & 11.1 \\
    CFNet\cite{CFNet} & 1.23 & 1.58 & 22.2 \\
    Abc-Net\cite{AbcNet} & 1.18 & 1.59 & {-} \\
    CAL-Net\cite{CALNet} & 1.19 & 1.53 & {-} \\
    AcfNet\cite{AcfNet} & 1.17 & 1.54 & \underline{5.5} \\
    ACVNet\cite{ACVNet} & 1.13 & \underline{1.47} & 7.1 \\
    IGEV-Stereo\cite{IGEV} & \underline{1.12} & \textbf{1.44} & 12.6 \\
    \hline
    DVANet(Ours) & \textbf{1.09} & 1.52 & \textbf{5.1} \\
    \hline
    \end{tabular}%
  \label{tab:sota_kitti12}%
\end{table}%

\textbf{KITTI 2012.} The model pretrained on Scene Flow is first fine-tuned on the combined KITTI 2012 and 2015 datasets for 20 epochs, and then solely on the KITTI 2012 for 10 epochs. To demonstrate the superiority of our model, we also compare the number of parameters as presented in Table \ref{tab:sota_kitti12}. Our DVANet achieves competitive results with less complexity on the KITTI 2012 leaderboard.

\subsection{Effectiveness of WRDE} \label{Effectiveness of WRDE}

\begin{figure}[t]
\setlength{\abovecaptionskip}{0.3cm}
\setlength{\belowcaptionskip}{-0.2cm}
  \centering
   \includegraphics[width=0.9\linewidth]
   {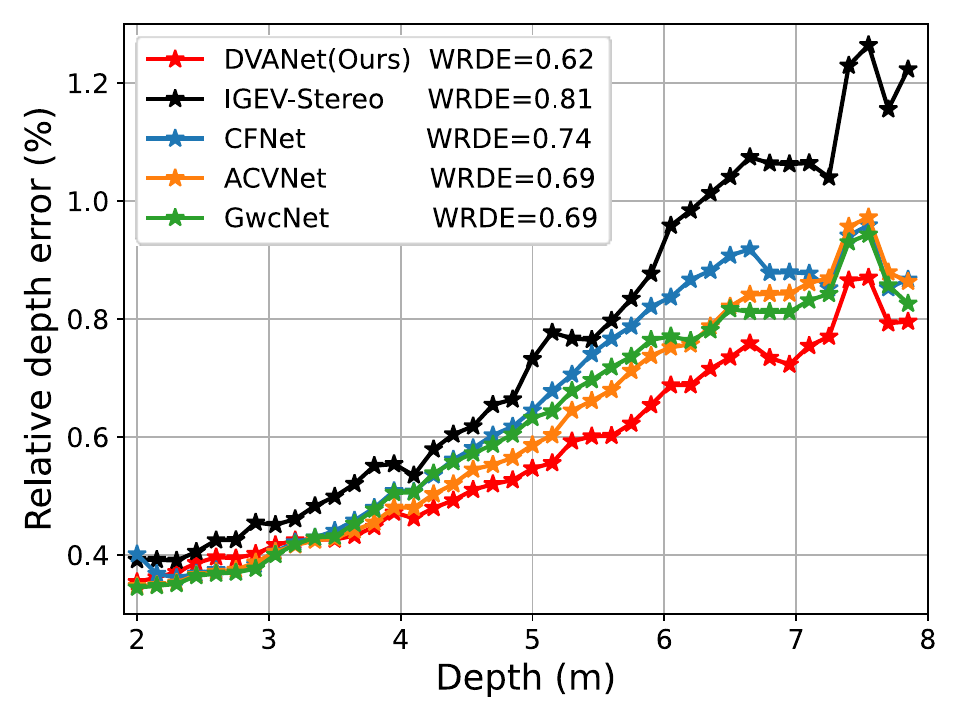}
   \vspace{-6pt}
   \caption{Comparison of relative depth errors on the RSRD.}
   \label{fig:wrde_rsrd}
\end{figure}

To validate the effectiveness of the proposed WRDE metric, we compare the relative depth errors of several models in Table \ref{tab:sota_rsrd}, as visualized in Figure \ref{fig:wrde_rsrd}. Table \ref{tab:wrde_segments} further lists their averaged relative errors within the three depth ranges. The error curves also present an increasing trend with respect to depth, reflecting the severe texture deterioration of road surface images in RSRD \cite{rsrd}. The trend of the WRDE values is consistent with the other metrics, i.e., lower EPE or 1-pixel error correspond to lower WRDE. 

\begin{table}[t]
  \centering
  \caption{The averaged relative depth errors within the near, medium, and far ranges in Figure \ref{fig:wrde_rsrd}. Different from the WRDE, the errors here are directly averaged without weighting.}
    \begin{tabular}{p{7.1em}|c|c|c}
    \hline
    Method & Near (\%) & Medium (\%) & Far (\%)\\
    \hline
    IGEV-Stereo\cite{IGEV} & 0.45 & 0.70 & 1.09\\
    CFNet\cite{CFNet} & 0.40 & 0.65 & 0.89 \\
    ACVNet\cite{ACVNet} & \textbf{0.39} & 0.59 & 0.85 \\
    GwcNet\cite{GwcNet} & \textbf{0.39} & 0.62 & 0.83 \\
    \hline
    DVANet(Ours) & 0.40 & \textbf{0.55} & \textbf{0.76} \\
    \hline
    \end{tabular}%
  \label{tab:wrde_segments}%
\end{table}%

The metric's usability and superiority are further validated through depth-wise performance assessment. Table \ref{tab:sota_rsrd} indicates that the EPE values of our DVANet and GwcNet \cite{GwcNet} differ by only 0.01, making it challenging to draw convincing conclusions based solely on this metric. However, the error curves in Figure \ref{fig:wrde_rsrd} imply that our DVANet significantly outperforms the other models, especially at medium and far ranges. Table \ref{tab:wrde_segments} further reinforces this observation, showing that DVANet achieves equivalent performance at the near range while outperforms with increasing depth, confirming the effectiveness of the proposed modules for texture-less matching. 

The insightful analysis method and metric provide a penetrative assessment of model performance, showing great potential for application in other depth estimation tasks. To use the metric, the depth range of interest should be first set. 
For better visualization and stable calculation, we recommend setting the depth interval between $3\%\sim10\%$ of the full depth range of interest. In this paper, we divide the depth range into three parts for prototype purpose. More segments and more sophisticated weighting strategies can be developed based on the distributions of target datasets.



\begin{table*}[ht]
  \centering
  \caption{Ablation studies. We compare two types of volume construction: correlation and discrepancy. \emph{Hire. Att.} means adopting the depth-aware texture hierarchy attention, while \emph{Disp. Att.} refers to the target-aware disparity attention module. }
    \begin{tabular}{cc|c|c||c|c|c|c|c}
    \hline
    \multicolumn{2}{c|}{{Volume}} & \multirow{2}{*}{\makecell[c]{Hier. \\ Att.}} & \multirow{2}{*}{\makecell[c]{Disp. \\ Att.}} & \multicolumn{3}{c|}{RSRD} & \multicolumn{2}{c}{Scene Flow}  \\
    \cline{1-2}  \cline{5-9}  \multicolumn{1}{c|}{Corr.} & \multicolumn{1}{c|}{Discr.} & & & EPE(px) & \textgreater 1px (\%) & WRDE (\%) & EPE(px) & \textgreater 1px (\%) \\
    \hline
    $\checkmark$ &   &   &   & 0.174 & 0.424 & 0.76 & - & - \\
    $\checkmark$ &   & $\checkmark$  &   & 0.164 & 0.368 & 0.69 & - & - \\
    $\checkmark$ &   &   & $\checkmark$ & 0.161 & 0.352 & 0.68 & - & - \\
    \hline
   & $\checkmark$  &   &   & 0.165 & 0.400 & 0.73 & 0.759 & 7.270 \\
   & $\checkmark$  &   & $\checkmark$  & 0.160 & 0.366 & 0.66 & 0.711 & 6.736 \\
   & $\checkmark$  & $\checkmark$  & $\checkmark$  & \textbf{0.154} & \textbf{0.343} & \textbf{0.62} & \textbf{0.697} & \textbf{6.510} \\
   \hline
  \end{tabular}%
  \label{tab:ablation}%
\end{table*}%


\subsection{Ablation Studies} \label{Ablation studies}

As shown in Table \ref{tab:ablation}, we conduct various ablation studies to demonstrate the effectiveness of the proposed modules. Experiments are performed on the RSRD \cite{rsrd} and Scene Flow \cite{DispNet} datasets for 5 epochs. To showcase the superiority of the discrepancy volume for texture-less matching, we compare it with the correlation operation. The configuration in the first row employs only the correlation volume, resulting in a model similar to GwcNet \cite{GwcNet}. Despite the reduction, it still outperforms GwcNet \cite{GwcNet} in terms of 1-pixel error rate, confirming the effectiveness of the feature extraction backbone. Results in the second row indicate significant improvements after introducing the texture hierarchy attention module. By focusing on target similarity features at the output stage, the disparity attention mechanism prominently reduces the 1-pixel outlier by 17.0\%.

Results in the last three rows illustrate that the discrepancy volume is more suitable for texture-less images, showing a significant improvement by solely replacing the correlation operation with subtraction when constructing the similarity volume. Furthermore, the texture hierarchy attention performs better when collaborating with the discrepancy volume compared to the correlation volume. The potential reason is that the subtraction operation provides more effective gradient information to both the feature extraction and depth estimation branches. It is worth noting that the proposed modules bring more enhancement on the RSRD dataset than on Scene Flow. This difference is attributed to the fact that texture deterioration in synthetic images is not as severe as in natural scenarios. Our DVANet proves more capable of capturing texture hierarchies in practical outdoor images.



\section{Conclusion}

We propose DVANet, an inspirational and light-weight scheme for practical texture-less stereo matching. The depth-aware texture hierarchy attention module functions by filtering the depth-related texture features. The target-aware disparity attention contributes to efficient aggregation with less computation. The developed strategies effectively suppress the performance corruption led by texture deterioration at far distances. The proposed WRDE metric for universal depth estimation tasks provides thorough performance indications. Our model achieves state-of-the-art performance on RSRD and competitive results on Scene Flow and KITTI datasets.

{
    \small
    \bibliographystyle{ieeenat_fullname}
    \bibliography{main}
}


\end{document}